\documentclass[11pt,a4paper]{article}
\usepackage[hyperref]{acl2018}
\usepackage{times}
\usepackage{latexsym}

\usepackage{amsmath,amssymb}
\usepackage{graphicx}

\usepackage{float}
\usepackage{footnote}
\usepackage{url}
\usepackage[utf8]{inputenc}

\usepackage[normalem]{ulem}
\definecolor{darkgreen}{rgb}{0.0, 0.5, 0.0}

\def\ODdel#1{\bgroup\markoverwith{\textcolor{darkgreen}{\rule[0.5ex]{2pt}{1pt}}}\ULon{#1}}

\def\YKdel#1{\bgroup\markoverwith{\textcolor{blue}{\rule[0.5ex]{2pt}{1pt}}}\ULon{#1}}

\def\VRdel#1{\bgroup\markoverwith{\textcolor{magenta}{\rule[0.5ex]{2pt}{1pt}}}\ULon{#1}}

\aclfinalcopy % Uncomment this line for the final submission
\title{Improving Context Modelling in Multimodal Dialogue Generation}

\author{Shubham Agarwal,$^\ast$\ \ Ondřej Dušek, Ioannis Konstas and Verena Rieser \\ 
The Interaction Lab,
Department of Computer Science\\
Heriot-Watt University, 
Edinburgh, UK\\
$^\ast$Adeptmind Scholar, Adeptmind Inc., Toronto, Canada\\
\texttt{\{sa201, o.dusek, i.konstas, v.t.rieser\}@hw.ac.uk} 
}

\date{}

\begin{document}
\maketitle
\begin{abstract} 
In this work, we investigate the task of textual response generation in a multimodal task-oriented dialogue system. Our work is based on the recently released Multimodal Dialogue (MMD) dataset \cite{saha2017multimodal} in the fashion domain.  We introduce a multimodal extension to the Hierarchical Recurrent Encoder-Decoder (HRED) model and show that this extension outperforms strong baselines in terms of text-based similarity metrics. %, as well as dialogue quality. 
We also showcase the shortcomings of current vision and language models by performing an error analysis on our system's output. 
%\keywords{Deep Learning, NLP, Computer Vision, Conversational AI, e-Commerce, Multimodal dialogue}
\end{abstract}

\section{Introduction}

This work aims to learn strategies for textual response generation in a multimodal conversation directly from data. Conversational AI has great potential for online retail: It greatly enhances user experience and in turn directly affects user retention \cite{chai2001natural}, especially if the interaction is multi-modal in nature. So far, most conversational agents are uni-modal -- ranging from open-domain conversation \cite{ram2018conversational,papaioannou2017alana,fang2017sounding} to task oriented dialogue systems \cite{rieser2010natural,rieser2011reinforcement,young2013pomdp,singh2000reinforcement,wen2016network}. While recent progress in deep learning has unified research at the intersection of vision and language, the availability of open-source multimodal dialogue datasets still remains a bottleneck.  
%\cite{xu2015show,vinyals2015show,karpathy2015deep,donahue2015long,ferraro2016visual,huang2016visual,antol2015vqa,lu2016hierarchical,malinowski2015ask,ren2015exploring,agrawal2016analyzing,das2017human,xu2016ask,tapaswi2016movieqa,zeng2017leveraging,tu2014joint,venugopalan2014translating,venugopalan2015sequence},

\begin{figure}[ht]
\centering
\includegraphics[scale=0.32]{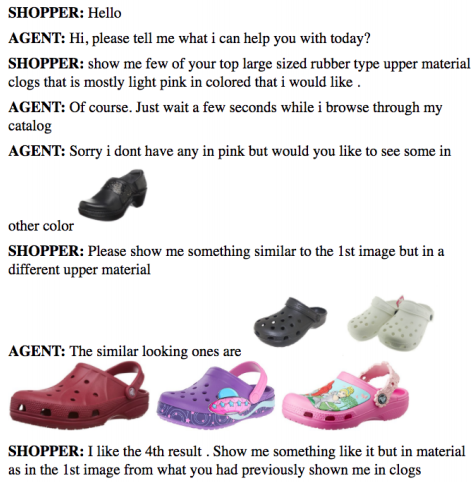} 
\caption{Example of a user-agent interaction in the fashion domain. 
In this work, we are interested in the textual response generation for a user query. Both user query and agent response can be multimodal in nature.}
\label{fig:test}
\end{figure}

This research makes use of a recently released Multimodal Dialogue (MMD) dataset \cite{saha2017multimodal}, which contains multiple dialogue sessions in the fashion domain. The MMD dataset provides an interesting new challenge, combining recent efforts on task-oriented dialogue systems, as well as visually grounded dialogue. In contrast to simple QA tasks in visually grounded dialogue, e.g.\ \cite{antol2015vqa}, it contains conversations with a clear end-goal. However, in contrast to previous slot-filling dialogue systems, e.g. \cite{rieser2011reinforcement,young2013pomdp}, it heavily relies on the extra visual modality to drive the conversation forward (see Figure~\ref{fig:test}).

In the following, we propose a fully data-driven response generation model for this task. Our work is able to ground the system's textual response with language and images by learning the semantic correspondence between them while modelling long-term dialogue context. 

\section{Model: Multimodal HRED over multiple images}
\label{sect:model}

Our model is an extension of the recently introduced Hierarchical Recurrent Encoder Decoder (HRED) architecture \cite{serban2016building,serban2017hierarchical,lu2016hierarchical}. In contrast to  standard  sequence-to-sequence models \cite{cho2014learning,sutskever2014sequence,bahdanau2014neural}, HREDs model the dialogue context by introducing a context Recurrent Neural Network (RNN) over the encoder RNN, thus forming a hierarchical encoder.

\begin{figure*}[ht]
 \centering
\includegraphics[scale=0.45]{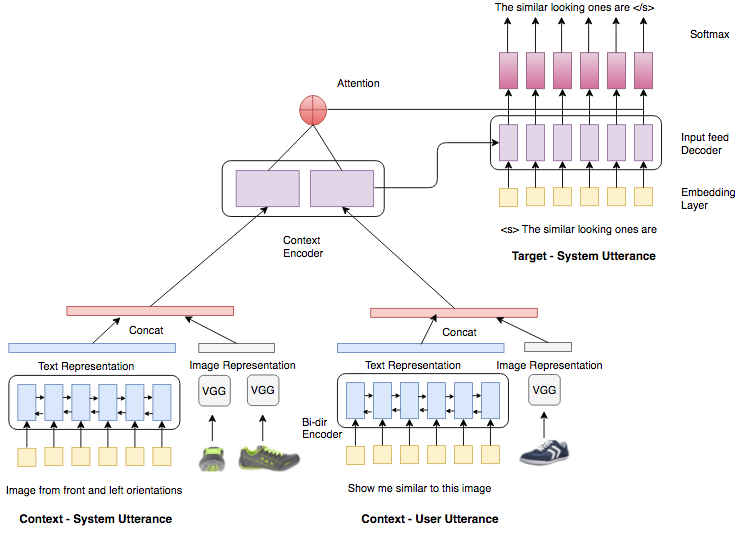}
\vspace{-3mm}
\caption{The Multimodal HRED architecture consists of four modules: utterance encoder, image encoder, context encoder and decoder. While \newcite{saha2017multimodal} `rollout' images to encode only one image per context, we concatenate all the `local' representations to form a `global' image representation per turn. Next, we concatenate the encoded text representation and finally everything gets fed to the context encoder. }
% The final hidden state of context RNN acts as initial hidden state of the decoder which generates output word by word conditioned on previous output token.
\label{fig:model}
\end{figure*}

We build on top of the HRED architecture to include multimodality over multiple images. A simple HRED consists of three RNN modules: encoder, context and decoder. In multimodal HRED, we combine the output representations from the utterance encoder with concatenated multiple image representations and pass them as input to the context encoder (see Figure~\ref{fig:model}). A dialogue is modelled as a sequence of utterances (turns), which in turn are modelled as sequences of words and images. Formally, a dialogue is generated according to the following: 
\vspace{-3mm}
\begin{equation}
\begin{split}
P_{\theta} (t_{1}, \dots t_{N}) & = \prod_{n=1}^{N} P_{\theta} (t_{n} | t_{<n} ) \\
\end{split}
\end{equation}
where $t_n$ is the $n$-th utterance in a dialogue. For each $m = 1, \dots , M_{n}$, we have hidden states of each module defined as:
\vspace{-1mm}
\begin{align}
h_{n,m}^{text} &= f_{\theta}^{text}(h_{n,m-1}^{text}, w_{m,n}) \\
h_{n}^{img} &= l^{img} ([g_{\theta}^{enc}(img_{1}), \dots g_{\theta}^{enc}(img_{k})]) \\
h_{n}^{cxt} &= f_{\theta}^{cxt}(h_{n-1}^{cxt}, [h_{n,M_{n}}^{text}, h_{n}^{img}]) \\
h_{n,m}^{dec}&=f_{\theta}^{dec}(h_{n,m-1}^{dec}, w_{n,m} ,h_{n-1}^{cxt})
\end{align}
\vspace{-10mm}
\begin{align}
h_{n,0}^{text} = 0 ;\quad h_{0}^{cxt} = 0; \quad h_{n,0}^{dec}=h_{N}^{cxt} 
\end{align}
where $f_{\theta}^{text}$,$f_{\theta}^{cxt}$ and $f_{\theta}^{dec}$ are GRU cells \cite{cho2014learning}. $\theta$ represent model parameters, $w_{n,m}$ is the $m$-th word in the $n$-th utterance and $g_{\theta}^{enc}$ is a Convolutional Neural Network (CNN); here we use VGGnet \cite{simonyan2014very}. We pass multiple images in a context through the CNN in order to get encoded image representations $g_{\theta}^{enc}(img_{k})$. Then these are combined together and passed through a linear layer $l^{img}$ to get the aggregated image representation for one turn of context, denoted by $h_{n}^{img}$ above. The textual representation $h_{n,M_{n}}^{text}$ is given by the encoder RNN $f_{\theta}^{text}$. Both $h_{n,M_{n}}^{text}$ and $h_{n}^{img}$ are subsequently concatenated and passed as input to the context RNN. $h_{N}^{cxt}$, the final hidden state of the context RNN, acts as the initial hidden state of the decoder RNN. Finally, output is generated by passing $h_{n,m}^{dec}$ through an affine transformation followed by a softmax activation. The model is trained using cross entropy on next-word prediction. During generation, the decoder conditions on the previous output token.

\begin{figure}[H]
%\vspace{-2mm}
\centering
\includegraphics[scale=0.47]{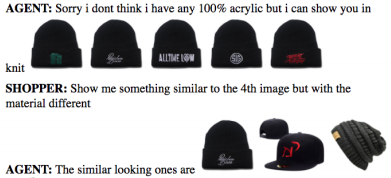}
\centering
%\resizebox{0.9\textwidth}{!}{
\scriptsize
\begin{tabular}{l}
  \hline
Our version of the dataset \\
  \hline 
  \textbf{Text Context:} Sorry i don't think i have any 100 \% acrylic but i can show  \\
you in knit $|$ Show me something similar to the 4th image but with the\\ material different \\
\textbf{Image Context:} [Img 1, Img 2, Img 3, Img 4, Img 5] $|$ [0, 0, 0, 0, 0] \\
\textbf{Target Response:} The similar looking ones are \\
 \hline 
%  \vspace{-1mm}\\
%  \hline
 Saha et al. \cite{saha2017multimodal} \\\hline 
\textbf{Text Context:} $|$ \\
\textbf{Image Context:}  Img 4 $|$ Img 5 \\
\textbf{Target Response:} The similar looking ones are \\\hline
\end{tabular}
\caption{Example contexts for a given system utterance; note the difference in our approach from \newcite{saha2017multimodal} when extracting the training data from the original chat logs. For simplicity, in this illustration we consider a context size of 2 previous utterances. `$|$' differentiates turns for a given context. We concatenate the representation vector of all images in one turn of a dialogue to form the image context. If there is no image in the utterance, we consider a $0_{4096}$ vector to form the image context. In this work, we focus only on the textual response of the agent.}
\label{fig:diff}
%\end{table}
\vspace{-2mm}
\end{figure}

\newcite{saha2017multimodal} propose a similar baseline model for the MMD dataset, extending HREDs to include the visual modality. However, for simplicity's sake, they `unroll' multiple images in a single utterance to include only one image per utterance. While computationally leaner, this approach ultimately loses the objective of capturing multimodality over the context of multiple images and text. In contrast, we combine all the image representations in the utterance using a linear layer. We argue that modelling all images is necessary to answer questions that address previous agent responses. For example in Figure~\ref{fig:diff}, when the user asks ``what about the 4th image?'', it is impossible to give a correct response without reasoning over all images in the previous response. In the following, we empirically show  that our extension leads to better results in terms of text-based similarity measures, as well as quality of generated dialogues.

% \vspace{-2mm}
\section{Experiments and Results}
\label{sect:DatasetImplementationMetrics}
\subsection{Dataset}

The MMD dataset \cite{saha2017multimodal} consists of 100/11/11k train/validation/test chat sessions comprising 3.5M context-response pairs for the model. Each session contains an average of 40 dialogue turns (average of 8 words per textual response, 4 images per image response). The data contains complex user queries, which pose new challenges for multimodal, task-based dialogue, such as \textit{quantitative inference (sorting, counting and filtering)}: ``Show me more images of the 3rd product in some different directions'', \textit{inference using domain knowledge and long term context}: ``Will the 5th result go well with a large sized messenger bag?'', \textit{inference over aggregate of images:} ``List more in the upper material of the 5th image and style as the 3rd and the 5th'', \emph{co-reference resolution}.  Note that we started with the raw transcripts of dialogue sessions to create our own version of the dataset for the model. This is done since the authors originally consider each image as a different context, while we consider all the images in a single turn as one concatenated context (cf.~Figure~\ref{fig:diff}).

\subsection{Implementation}
We use the PyTorch\footnote{\url{https://pytorch.org/}} framework ~\cite{paszke2017automatic} for our implementation.\footnote{Our code is freely available at:\\ \url{https://github.com/shubhamagarwal92/mmd}} We used 512 as the word embedding size as well as hidden dimension for all the RNNs using GRUs~\cite{cho2014learning} with tied embeddings for the (bi-directional) encoder and decoder. The decoder uses Luong-style attention mechanism~\cite{luong2015effective} with input feeding. We trained our model with the Adam optimizer~\cite{kingma2014adam}, with a learning rate of 0.0004 and clipping gradient norm over 5. We perform early stopping by monitoring validation loss. For image representations, we use the FC6 layer representations of the VGG-19 \cite{simonyan2014very}, pre-trained on ImageNet.\footnote{In future, we plan to exploit state-of-the-art frameworks such as ResNet or DenseNet and fine tune the image encoder jointly, during the training of the model.} 
%Our vocabulary size is 7452. %(we apply a threshold of 2).

\begin{figure*}[ht]
\centering
\includegraphics[width=\textwidth]{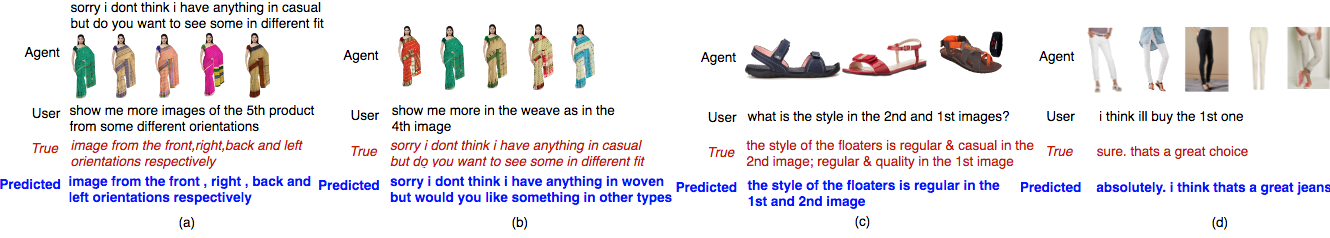} 
\vspace{-7mm}
\caption{Examples of predictions using M-HRED--attn (5). Recall, we are focusing on generating textual responses. Our model predictions are shown in blue while the true gold target in red. We are showing only the previous user utterance for brevity's sake.
}
\label{fig:results_predict}
\end{figure*}

\subsection{Analysis and Results}

We report sentence-level \textsc{Bleu-4} \cite{papineni2002bleu}, \textsc{Meteor} \cite{lavie2007meteor} and \textsc{Rouge-L} \cite{lin2004automatic} using the evaluation scripts provided by \cite{sharma2017nlgeval}. We compare our results against \newcite{saha2017multimodal} by using their code and data-generation scripts.\footnote{\url{https://github.com/amritasaha1812/MMD_Code}} Note that the results reported in their paper are on a different version of the corpus, hence not directly comparable. 

\begin{table}[ht]
\centering
% \small
\resizebox{0.48\textwidth}{!}{
\begin{tabular}{l|c|cccccc}
  \hline
Model & Cxt & \textsc{Bleu-4} &\textsc{Meteor} & \textsc{Rouge-L} \\
  \hline \hline
Saha et al. M-HRED* & 2 & 0.3767 & 0.2847 & 0.6235  \\
T-HRED & 2 & 0.4292 & 0.3269 & 0.6692\\
M-HRED & 2 & 0.4308 & 0.3288 & 0.6700\\
T-HRED--attn & 2 & 0.4331 & 0.3298 & 0.6710\\
M-HRED--attn & 2 &  0.4345 & 0.3315 & 0.6712\\
T-HRED--attn & 5 & 0.4442 & \textbf{0.3374} & 0.6797\\
\textbf{M-HRED--attn} & 5  & \textbf{0.4451} & 0.3371 & \textbf{0.6799} \\
\hline
\end{tabular}
}
\vspace{-0.5mm}
\caption{Sentence-level \textsc{Bleu-4}, \textsc{METEOR} and \textsc{ROUGE-L} results for the response generation task on the MMD corpus. ``Cxt'' represents context size considered by the model. Our best performing model is M-HRED--attn over a context of 5 turns. *Saha et al. has been trained on a different version of the dataset.}
\label{table:our_results}
\end{table}

Table \ref{table:our_results} provides results for different configurations of our model (``T'' stands for text-only in the encoder, ``M'' for multimodal, and ``attn'' for using attention in the decoder). We experimented with different context sizes and found that output quality improved with increased context size (models with 5-turn context perform better than those with a 2-turn context), confirming the observation by \newcite{serban2016building,serban2017hierarchical}.\footnote{Using pairwise bootstrap resampling test \cite{koehn2004statistical}, we confirmed that the difference of M-HRED-attn (5) vs. M-HRED-attn (2) is statistically significant at 95\% confidence level.} Using attention clearly helps: even T-HRED--attn outperforms M-HRED (without attention) for the same context size. We also tested whether multimodal input has an impact on the generated outputs. However, there was only a slight increase in BLEU score (M-HRED--attn vs T-HRED--attn). 

To summarize, our best performing model (M-HRED--attn) outperforms the model of Saha et al. by 7 BLEU points.\footnote{The difference is statistically significant at 95\% confidence level according to the pairwise bootstrap resampling test \cite{koehn2004statistical}.} This can be primarily attributed to the way we created the input for our model from raw chat logs, as well as incorporating more information during decoding via attention. Figure \ref{fig:results_predict} provides example output utterances using M-HRED--attn with a context size of 5. Our model is able to accurately map the response to previous textual context turns as shown in (a) and (c). In (c), it is able to capture that the user is asking about the style in the 1st and 2nd image. (d) shows an example where our model is able to relate that the corresponding product is `jeans' from visual features, while it is not able to model fine-grained details like in (b) that the style is `casual fit' but resorts to `woven'. 
% Similarly, in (c) the model is able to predict coarse style types as `regular' in both images while it cannot tell between the more fine-grained types of `regular casual' and `regular quality', indicating limitations in the corresponding image representations to provide these distinguishing details between the same kind of images. 

\section{Conclusion and Future Work}
\label{sect:ConclusionFutureWork}

In this research, we address the novel task of response generation in search-based multimodal dialogue by learning from the recently released Multimodal Dialogue (MMD) dataset \cite{saha2017multimodal}. We introduce a novel extension to the Hierarchical Recurrent Encoder-Decoder (HRED) model \cite{serban2016building}
 and show that our implementation significantly outperforms the model of~\citet{saha2017multimodal} by modelling the full multimodal context. Contrary to their results, our generation outputs improved by adding attention and increasing context size. However, we also show that multimodal HRED does not improve significantly over text-only HRED, similar to observations by \newcite{agrawal2016analyzing} and \citet{qian2018multimodal}.
 Our model learns to handle textual correspondence between the questions and answers, while mostly ignoring the visual context. This indicates that we need better visual models to encode the image representations when he have multiple similar-looking images, e.g., black hats in Figure~\ref{fig:diff}. We believe that the results should improve with a jointly trained or fine-tuned CNN for generating the image representations, which we plan to implement in future work.
 
% \vspace{-10mm}
 
\section*{Acknowledgments}
This research received funding from Adeptmind Inc., Toronto, Canada and the MaDrIgAL EPSRC project
(EP/N017536/1). The Titan Xp used for this
work was donated by the NVIDIA Corp.

%
%The acknowledgments should go immediately before the references.  Do not number the acknowledgments section ({\em i.e.}, use \verb|\section*| instead of \verb|\section|). Do not include this section when submitting your paper for review.
% \newpage
%\bibliography{acl2018}
\bibliography{biblio}
\bibliographystyle{acl_natbib}

%\appendix
%\section{Appendix}
%
%\begin{figure}[ht]
%\centering{\includegraphics[scale=.3]{pics/data_v2}}
%\caption{Snippet of a chat log in the MMD dataset. Shopper comes on the platform and provides her requirements in a dialogue setting to retrieve a relevant product.}
%\label{fig:data_v2}
% \end{figure}

\end{document}